\documentclass[11pt]{article}

\usepackage[final]{acl}

\usepackage{times}
\usepackage{latexsym}

\usepackage[linesnumbered,ruled,vlined]{algorithm2e}

\usepackage{amsmath, amssymb}
\usepackage{hyperref}
\usepackage{booktabs}
\usepackage{adjustbox}
\usepackage{tcolorbox}

\usepackage{booktabs}     
\usepackage{amssymb}      
\usepackage{makecell}     
\usepackage{array}        
\usepackage{stfloats}
\tcbuselibrary{listings}

\usepackage[T1]{fontenc}

\usepackage[utf8]{inputenc}

\usepackage{microtype}

\usepackage{inconsolata}
\usepackage{subcaption}
\usepackage{graphicx}
\usepackage{float}
\makeatletter
\DeclareRobustCommand{\printfontsize}{\f@size}
\makeatother

\title{Sparrow: Text-Anchored Window Attention with Visual-Semantic Glimpsing for Speculative Decoding in Video LLMs}

\author{
    \textbf{Libo Zhang} \quad \textbf{Zhaoning Zhang$ $\thanks{\ \ indicates corresponding authors.}} \quad \textbf{Wangyang Hong} 
    \\
     \quad \textbf{Peng Qiao}  \quad \textbf{Dongsheng Li} \\[1em]
    National Key Laboratory of Parallel and Distributed Computing \\ 
    College of Computer Science and Technology \\
    National University of Defense Technology, Changsha, China.
    \\
 \texttt{ \{zhanglibo, zhangzhaoning, hongwanyang9527, } \\
 \texttt{pengqiao, dsli\}@nudt.edu.cn}
}

\begin{document}
\maketitle 
\begin{abstract}

Although speculative decoding is widely used to accelerate Vision-Language Models (VLMs) inference, it faces severe performance collapse when applied to Video Large Language Models (Vid-LLMs). The draft model typically falls into the trap of attention dilution and negative visual gain due to key-value cache explosion and context window mismatches. We observe a visual semantic internalization phenomenon in Vid-LLMs, indicating that critical visual semantics are implicitly encoded into text hidden states during deep-layer interactions, which renders raw visual inputs structurally redundant during deep inference. 
To address this, we propose the Sparrow\footnote{\url{https://github.com/ddInference/Sparrow}} framework, which first utilizes visually-aware text-anchored window attention via hidden state reuse to fully offload visual computation to the target model, and leverages intermediate-layer visual state bridging to train the draft model with semantic-rich intermediate states, thereby filtering out low-level visual noise. Additionally, a multi-token prediction strategy is introduced to bridge the training-inference distribution shift. Experiments show that Sparrow achieves an average speedup of 2.82× even with 25k visual tokens, effectively resolving the performance degradation in long sequences and offering a practical solution for real-time long video tasks.
\end{abstract}

\section{Introduction}
\label{sec:intro}

Video Large Language Models (Vid-LLMs) have achieved remarkable breakthroughs in spatiotemporal understanding and generation tasks ~\cite{li2024llava, bai2025qwen2, maaz2024video, zhang2024video}. However, this leap in capability incurs substantial computational costs: the rich spatiotemporal semantics inherent in long videos are mapped into extremely long input sequences, resulting in prohibitive GPU memory consumption and computational overhead. The inference latency bottleneck of current Vid-LLMs stems primarily from the dual constraints imposed by the generation paradigm and input representation. At the generation level, mainstream autoregressive decoding is constrained by the memory wall, where its token-by-token sequential mechanism imposes an inherent speed bottleneck ~\cite{fu2024break}. At the representation level, video data, once encoded ~\cite{wang2025internvideo2,zhang2023video}, produces massive visual tokens. This not only significantly exacerbates forward computational loads but also precipitates a rapid expansion of the key-value (KV) cache during long-sequence generation, posing a critical obstacle to large-scale deployment.
\begin{figure}[t]
  \vspace{1cm}
  \centering
  \includegraphics[width=\linewidth]
  {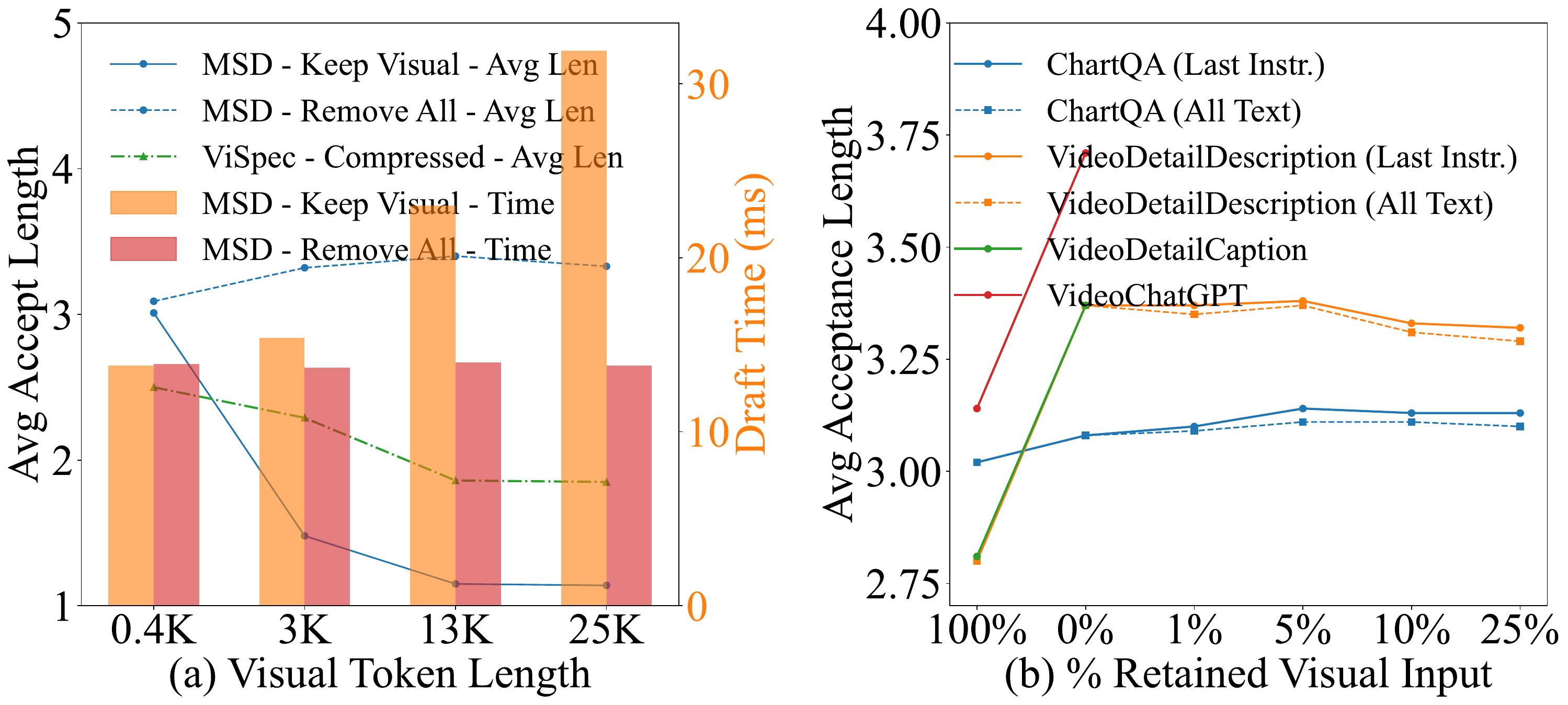}
  \caption{Analysis of the impact of visual token length on performance. (a) Comparison of average accepted length and drafting latency between MSD and ViSpec on the VideoDetailCaption dataset across varying visual sequence lengths. (b) Performance on image and video tasks with MSD at different visual retention rates. The Last Instr. and All Text strategies retain the top-x\% visual tokens based on the attention scores of the final instruction token and the average attention scores of all text tokens towards visual regions, respectively. ViSpec and MSD employ Qwen2.5-VL-7B and Qwen2-VL-7B as target models, respectively, with both utilizing the official released weights for their draft models.}
  \label{fig:ccd_example_1}
\end{figure}
To mitigate this bottleneck, previous studies have explored multi-granular visual pruning strategies ~\cite{chen2024image,liu2024multi,xing2024pyramiddrop,tao2025dycoke,zhang2025vscan}, premised on the prior that visual tokens exhibit significant redundancy. Although such methods reduce memory usage during the prefill stage, they inherently constitute lossy compression: the direct removal of input information inevitably compromises the model's fine-grained perception capabilities. In contrast, speculative decoding~\cite{hu2025speculative,chen2023accelerating,leviathan2023fast} provides a superior, lossless alternative by leveraging a lightweight draft model for rapid prediction followed by parallel verification by the target model.

However, adapting speculative decoding to long-video scenarios presents formidable challenges. Existing multimodal speculative decoding, primarily tailored for short-sequence image tasks, suffer from a precipitous decline in acceleration gains when applied to long videos (as illustrated in Figure~\ref{fig:ccd_example_1}, the average accepted lengths of MSD ~\cite{lin2025speculative} and ViSpec ~\cite{kang2025ViSpec} plummet by 63\% and 30\%, respectively). This performance collapse stems from two critical obstacles. First is the memory access bottleneck within the draft model. The KV cache explosion induced by long videos significantly increases the inference latency of the lightweight model, effectively negating the temporal dividends gained from speculation. Second is the inherent conflict between model architecture and sequence length. On one hand, long video sequences (often exceeding 10k tokens) far surpass the pre-training context windows of mainstream lightweight models (typically capped at ~2,048 tokens), leading to length mismatch and forced information truncation. On the other hand, excessively long sequences of visual tokens trigger severe attention dilution. The naïve application of existing multimodal speculative decoding methods leaves capacity-constrained draft models struggling to extract critical features from massive visual inputs, ultimately resulting in a drastic degradation of speculative accuracy and overall efficiency.

Investigating the root causes of this phenomenon, we identify a critical factor in long-video scenarios: the negative gain of visual information. For capacity-constrained draft models, retaining a massive number of visual tokens proves more detrimental than beneficial; the tens of thousands of visual features effectively manifest as computational noise, causing speculative performance to degrade significantly as the visual sequence length increases. 
Concurrently, an analysis of the information flow within Vid-LLMs reveals a pivotal mechanism: visual semantic internalization. Through multi-stage inter-layer interactions, Vid-LLMs implicitly internalize essential visual semantics, encoding them directly into the hidden states corresponding to text positions. Consequently, during the deep inference stage, the explicit stream of raw visual inputs becomes structurally redundant.


Based on these insights, we propose the visually-aware text-anchored window attention via hidden state reuse (HSR-VATA) strategy. The core of this method is computation offloading, aiming to transfer the intensive visual processing tasks entirely to the target model. We observe that the target model inherently integrates comprehensive visual information and encodes it into text hidden states; therefore, the draft model need only reuse these ready-made states via the HSR mechanism. This reuse acts as an efficient glimpse into the massive visual stream, enabling the draft model to rapidly capture key semantics and avoid the high overhead of processing raw visual sequences. Coupled with VATA to confine attention to text anchor positions, this framework allows the draft model to maintain its visual perception capabilities without explicitly receiving or processing raw visual tokens.


Furthermore, addressing the difficulty of the draft model's minimalist architecture in handling low-level raw visual noise, we propose intermediate-layer visual state bridging (IVSB) during the training phase. By directly extracting intermediate-layer visual states from the target model, specifically those where semantic interaction is most active and alignment is implicitly established, as the visual information source for training the draft model, we leverage the multi-stage processing characteristics of Vid-LLMs. This approach preserves high-level semantic features while effectively filtering out low-level redundancy and noise from raw visual signals, achieving efficient cross-modal alignment during training. Simultaneously, by incorporating a multi-token prediction strategy, we effectively bridge the distribution discrepancy between training and inference.

\paragraph{Contributions.} The main contributions of this paper are as follows:
\begin{itemize}
    \item To the best of our knowledge, this is the first work to apply a lightweight draft model to Vid-LLMs. We reveal attention dilution and negative gain in long-video speculative decoding, validating visual offloading feasibility via the target model's deep hidden states.
    \item We propose the Sparrow framework. By integrating HSR-VATA, we eliminate visual redundancy and attention dilution. Furthermore, by combining IVSB with multi-token prediction, we filter visual noise while reinforcing cross-modal alignment, effectively resolving the training-inference distribution shift.
    \item Experiments demonstrate that even with ultra-long visual inputs of 25k tokens, our method achieves an average speedup of 2.82×, while effectively overcoming the bottleneck of performance degradation in long-video scenarios.
\end{itemize}

\begin{figure}[t]
  \centering
  \includegraphics[width=\linewidth]{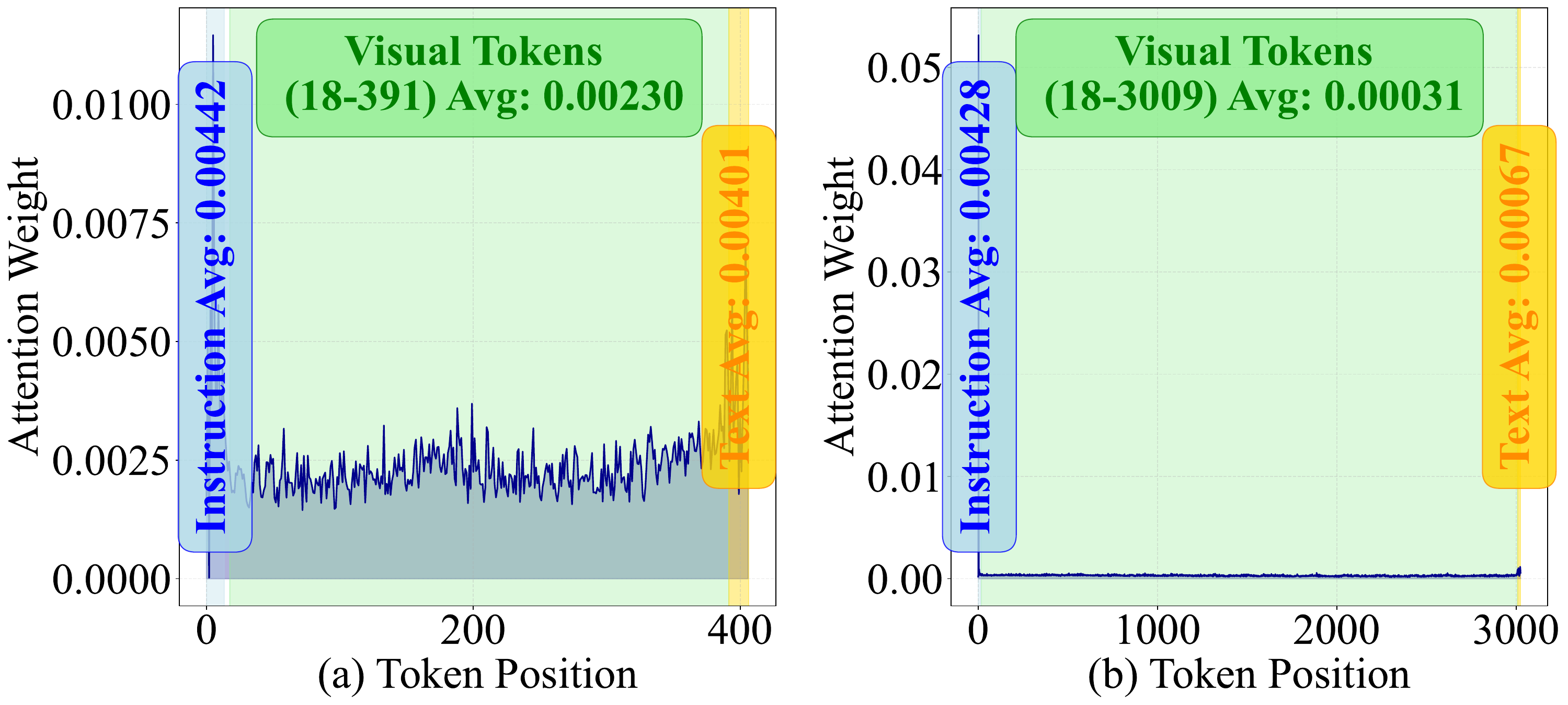}
  \caption{(a) \& (b) Visualization of the average attention weight distribution from the last instruction token to preceding Instruction, Visual, and Text tokens in short (0.4k) and long (3k) sequence tasks.}
  \label{fig:architecture}
\end{figure}

\section{Insight }
\label{sec:Insight}
Current research on multimodal speculative decoding ~\cite{hu2025dream,huo2025spec,lin2025speculative,leebatch,ganesan2025massv,kang2025ViSpec} primarily focuses on image tasks, where draft models typically employ either full (e.g., MSD) or compressed (e.g., ViSpec) visual information as input. In LLaVA~\cite{liu2024improved}, for instance, visual tokens typically number around 576, accounting for merely 64\% of the total input sequence ~\cite{chen2024image}. However, as the domain transitions towards long video tasks, the input distribution shifts drastically. Taking Qwen2.5-VL-7B as an example, video encoding can generate up to 25k visual tokens, with their proportion surging to over 99\%. This severe distributional shift compels us to re-examine a critical question: Can parameter-constrained draft models effectively process such a massive volume of visual information?

\subsection{Performance Bottlenecks in Long-Video Scenarios}
 
We first investigate the scalability of existing strategies in long video scenarios. As shown in Figure \ref{fig:ccd_example_1} (a), current methods encounter severe bottlenecks as sequence length grows. The full-input strategy (e.g., MSD) suffers a critical efficiency regression: as video token length expands from 0.4k to 25k, the average accepted length decays by 63\% while latency doubles, resulting in a negative speedup. Meanwhile, compression-based strategies (e.g., ViSpec), constrained by their training on short sequences, fail to capture complex spatiotemporal dynamics; this leads to a significant 30\% performance drop despite the reduced computational overhead.
\begin{figure}[t]
  \centering 
  \includegraphics[width=\linewidth]
  {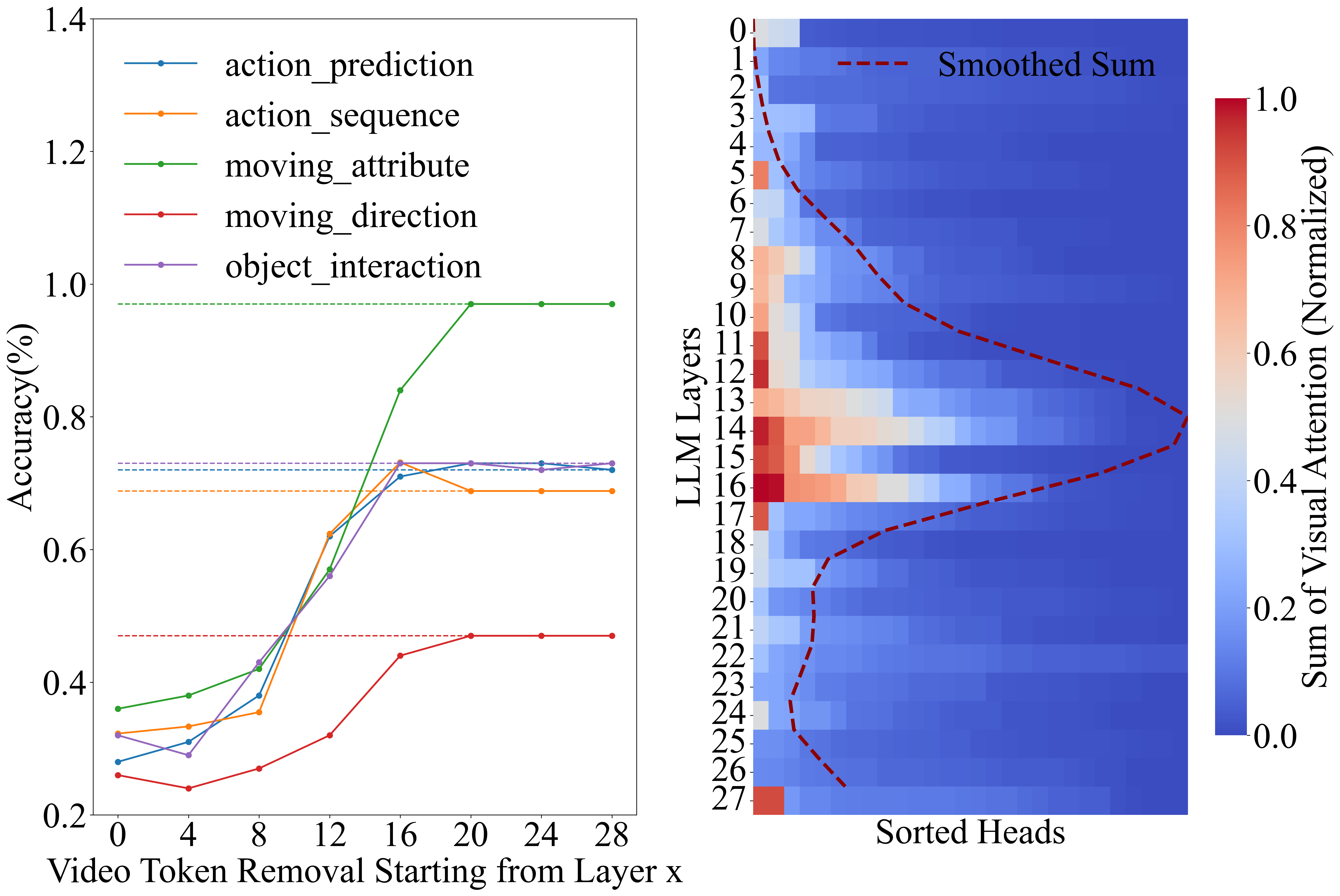}
  
  \caption{Layer-wise Importance Analysis of Visual Tokens in Qwen2.5-VL-7B. (a) Comparison of task accuracy (solid lines) after removing visual tokens starting from layer x versus the native baseline (dashed lines). (b) Sum of attention received by all visual tokens from the last instruction token across different attention heads and model layers.
}
  \label{fig:architecture1}
\end{figure}
\begin{figure*}[t]
  \centering
  \includegraphics[width=0.95\textwidth, keepaspectratio]{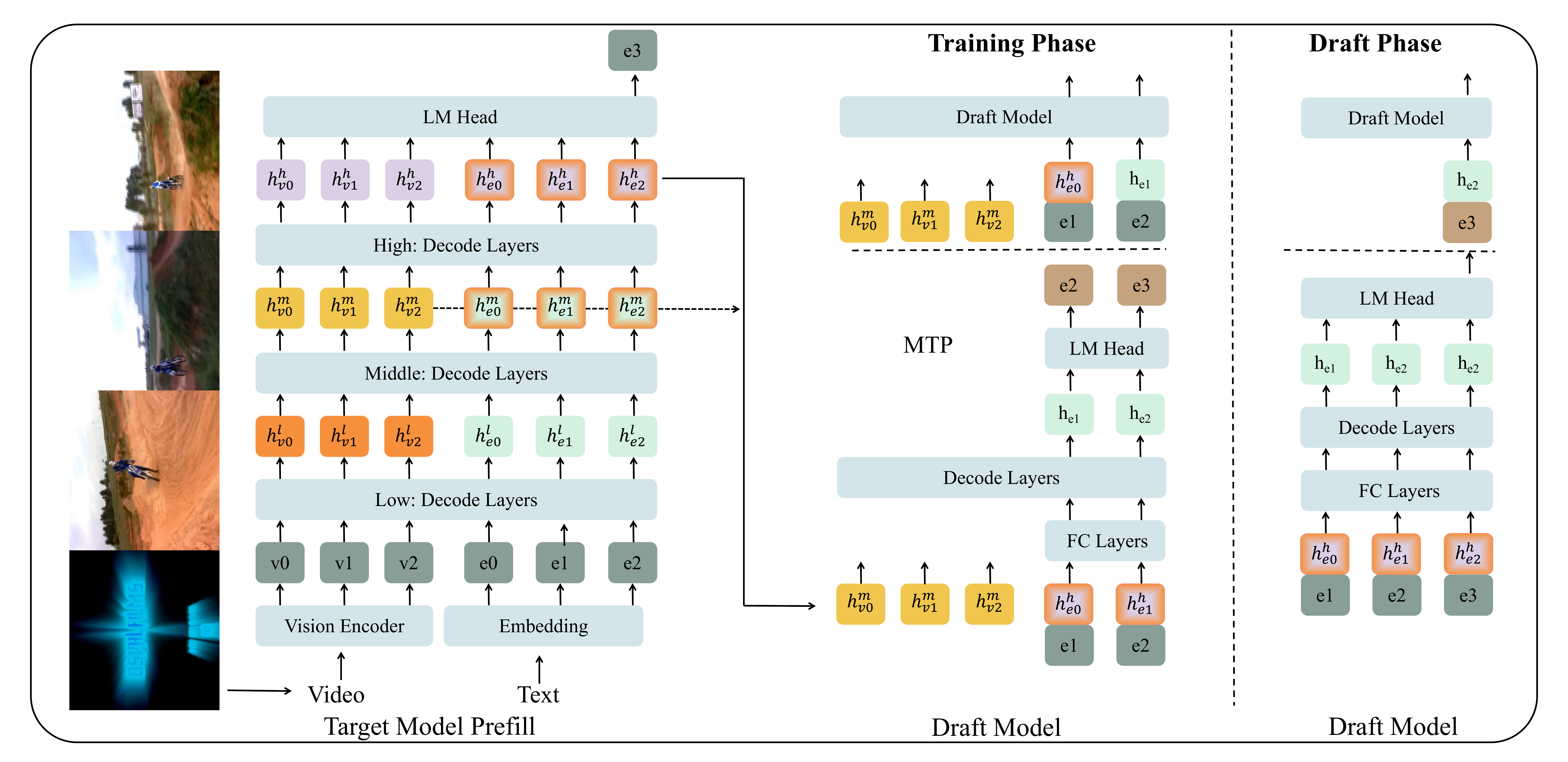} 
  
  \caption{Illustration of the Sparrow framework. The left and right panels illustrate the training and inference phases, respectively. Initially, the target model performs multi-stage fusion on visual embeddings $v_i$ and text embeddings $e_i$, yielding a noise-filtered visual hidden state $h^m_{v_i}$ and a visually-infused text hidden state $h^h_{e_i}$. In the training phase, $h^m_{v_i}$ and $h^h_{e_i}$ are concatenated and fed into the draft model to produce $h_{e_i}$. Subsequently, $h^m_{v_i}$ is concatenated with $h_{e_i}$ and re-input into the draft model. Losses from both stages are jointly computed to mitigate the discrepancy between training and inference. In the inference phase, the draft model directly takes $h^h_{e_i}$ as input.}
  
  \label{fig:architecture_3}
\end{figure*}
\subsection{Redundancy of Visual Information and Negative Gain Phenomenon} 
To unveil the true contribution of visual information during inference, we implemented multi-granularity visual token pruning based on attention distributions, contrasting the distinct response patterns between image and video tasks (see Figure \ref{fig:ccd_example_1} (a) \& (b) and Table \ref{tab:visual_ablation}). Experimental results indicate that the average accepted length in image tasks is insensitive to pruning ratios, suggesting a neutral characteristic of visual redundancy. Conversely, video tasks exhibit a counter-intuitive negative correlation: the average accepted length increases with fewer visual tokens. Furthermore, the model demonstrates exceptional robustness under zero-visual-input settings, remaining virtually unaffected by the original video token length (0.4k to 25k). This significant discrepancy provides compelling evidence that excessive video sequences in the drafting phase are not only highly redundant but effectively constitute noise that suppresses language generation.

\subsection{Attribution Analysis I: Attention Dilution Under Limited Capacity}   
Why does visual information transition into a detrimental factor in long videos? Integrating the analysis of Figure \ref{fig:architecture} (a) \& (b) , we attribute the root cause to the limited attention capacity of small-scale draft models. In short-sequence scenarios (images/short videos), while the performance stability upon removing visual inputs confirms high redundancy, this redundancy proves non-fatal: the model, leveraging the selectivity of self-attention, successfully anchors onto key regions (Figure \ref{fig:architecture} (a)) to automatically suppress interference. Conversely, the situation deteriorates drastically for long videos. Tens of thousands of visual tokens not only breach the effective boundaries of positional encoding but also precipitate severe attention dilution. Under the barrage of massive visual noise, the capacity-constrained model falls into cognitive overload, where attention resources are involuntarily scattered across irrelevant details, failing to distinguish valid information (Figure \ref{fig:architecture} (b)). This mechanistic failure directly undermines sequence modeling capability, resulting in a distinct monotonic decline in average accepted length as video token length increases (Figure \ref{fig:ccd_example_1} (a)).

\subsection{ Attribution Analysis II: Layer-wise Interaction and Deep Internalization of Visual Semantics} 
\label{sec:visual_internalization}

Draft models typically utilize the second-to-last layer hidden states of the target model as input. To investigate why stripping visual information during inference does not compromise performance, we analyzed the internal information flow via a layer-wise truncation experiment (as shown in  Figure \ref{fig:architecture1} (a)). The results reveal a distinct stage-wise dependency: while explicit visual signals are indispensable in the early network stages, the prediction performance remains robust beyond the 20th layer, even after completely removing the visual flow and relying solely on text hidden states. We hypothesize that as network depth increases, visual semantics are gradually fused and encoded into the text hidden states, rendering the original visual tokens a form of structural redundancy. This mechanism plausibly explains the phenomenon observed in Figure \ref{fig:ccd_example_1} (a), where physically removing visual input effectively eliminates redundancy-induced interference in high-noise scenarios such as long videos, thereby enhancing model performance.

Given the apparent low dependence of deep layers on explicit visual information, we further employed attention heatmaps to locate cross-modal interaction zones (as shown in Figure \ref{fig:architecture1} (b)), forming strong mutual corroboration with the aforementioned truncation experiment. Visual analysis indicates that the intermediate layers of the LLM serve as the core arena for visual-text interaction, whereas the shallow and deep layers focus predominately on text processing. This finding is highly consistent with the performance trajectory in Figure \ref{fig:architecture1} (a): the model's predictive capability rises sharply during the intense interactions of the middle layers and stabilizes synchronously as interaction intensity wanes in the deep layers. This correspondence further supports the hypothesis that visual information is integrated into text representations primarily within the intermediate stages.

\section{Method}

\begin{table}[t]
    \centering
    \caption{Impact of visual information injection on draft model performance during training.}
    \label{tab:visual_ablation_1}
    \begin{tabular}{lcc}
        \toprule
        Training \& Inference Input & ChartQA & AI2D \\
        \midrule
        Visual + Text         & 4.25 & 4.28 \\
        0\% Visual + Text     & 4.23 & 4.25 \\
        50\% Visual + Text    & 4.21 & 4.24 \\
        \bottomrule
    \end{tabular}
\end{table}

\subsection{Visually-Aware Text-Anchored Window Attention via Hidden State Reuse}
\label{sec:visual_internalization31}

Section \ref{sec:visual_internalization} reveals the phenomenon of visual semantic internalization : deep visual information is fused into text hidden states, turning raw visual signals into structural redundancy. Based on this, we propose the cross-modal task offloading strategy. This strategy completely offloads the expensive visual feature extraction and cross-modal alignment tasks to the target model, allowing the draft model to focus solely on prediction using text hidden states that carry visual context, thereby bypassing its structural limitations.

To this end, we design a hidden state reuse (HSR) mechanism to reconstruct the input. Instead of processing the original long sequence, the draft model receives an augmented input $z_t$. As illustrated in Figure \ref{fig:architecture_3}, we utilize a projection layer $FC$ to perform dimensionality reduction on the concatenated features of the current text embedding $e_t$ and the target model's text hidden state from the previous timestep $h^h_{e_{t-1}}$, which encapsulates the visual context:
\begin{equation}
    z_t = FC\left( e_t \oplus h^h_{e_{t-1}} \right)
\end{equation}

where $e_t$ provides explicit token identity to eliminate uncertainty ~\cite{li2024eagle}, while $h^h_{t-1}$ inherits cross-modal representations deeply integrated by the target model. This direct state reuse is akin to casting an efficient Glimpse at the vast stream of visual information. Through this glimpse, the draft model acquires relevant visual information, thereby achieving visual offloading at the physical level. This enables the model to circumvent redundant computations on the raw sequence while retaining the capacity to capture the necessary visual context. Note that the superscript $h$ denotes the second-to-last layer of the target model.

Given that $z_t$ has internalized visual semantics, we propose visually-aware text-anchored window attention (VATA). VATA discards the visual KV cache, forcing attention closure strictly within the text domain $\mathcal{T}$:

\begin{equation}
Attention_{VATA} = Softmax\left(\frac{Q_t K_{\mathcal{T}}^T}{\sqrt{d}}\right) V_{\mathcal{T}}
\end{equation}

Specifically, within the equation, $Q_t$ denotes the query vector at the current timestep $t$, and $d$ represents the hidden dimension used for scaling dot-product attention, while $K_{\mathcal{T}}$ and $V_{\mathcal{T}}$ correspond to the Key and Value matrices restricted strictly to the text domain $\mathcal{T}$. VATA is based on the premise of visual semantic internalization, limiting computation to text-anchored positions. Crucially, after HSR interaction, representations at these positions have evolved into visual-text fusion representations, rather than raw textual information. This design ensures the draft model infers directly based on deeply integrated cross-modal representations, drastically reducing complexity from $O((L_{vis}+L_{txt})^2)$ to the pure text level $O(L_{txt}^2)$, where $L_{vis}$ and $L_{txt}$ denote the sequence lengths of visual and textual tokens, respectively.

\begin{table*}[t]
  \centering
  \caption{We evaluate the performance of LLaVA-OneVision-7B on long-video benchmarks using NVIDIA L20 GPUs, and report the average accepted length ($\tau$), the end-to-end wall-clock speedup ratio (ESR), and the decoding speedup ratio (DSR) for the generation phase, with the visual token length fixed at 25K across all datasets.
}
  \label{tab:video_performance} 
  
  \begin{adjustbox}{width=\linewidth, center}
    \small 
    \begin{tabular}{l ccc ccc ccc ccc ccc}
      \toprule
      Method & \multicolumn{3}{c}{VideoDetailCaption} & \multicolumn{3}{c}{MVBench} & \multicolumn{3}{c}{LongVideoBench} & \multicolumn{3}{c}{VideoMME} & \multicolumn{3}{c}{Avg} \\
      
      \cmidrule(lr){2-4} \cmidrule(lr){5-7} \cmidrule(lr){8-10} \cmidrule(lr){11-13} \cmidrule(lr){14-16}
       &  $\tau$ & ESR & DSR &  $\tau$ & ESR & DSR &  $\tau$ & ESR & DSR &  $\tau$ & ESR & DSR &  $\tau$ & ESR & DSR \\
      \midrule
      
      \multicolumn{16}{c}{Temp = 0} \\
      \midrule
      
      SpecVLM & \textbf{4.15} & 1.25x & 1.47x & \textbf{4.20} & 1.26x & 1.43x & \textbf{4.04} & 1.23x & 1.39x & 3.91 & 1.21x & 1.34x & \textbf{4.07} & 1.24x & 1.41x \\
      MSD     & 2.69 & 1.47x & 1.76x & 2.65 & 1.43x & 1.71x & 2.63 & 1.43x & 1.71x & 2.68 & 1.49x & 1.77x & 2.66 & 1.46x & 1.74x \\
      Ours    & 3.88 & \textbf{1.96x} & \textbf{2.84x} & 3.67 & \textbf{1.83x} & \textbf{2.69x} & 3.78 & \textbf{1.90x} & \textbf{2.77x} & \textbf{3.98} & \textbf{2.02x} & \textbf{2.96x} & 3.83 & \textbf{1.93x} & \textbf{2.82x} \\
      
      \midrule
      \multicolumn{16}{c}{Temp = 1} \\
      \midrule
      
      MSD     & 2.20 & 1.27x & 1.43x & 2.31 & 1.31x & 1.50x & 2.17 & 1.28x & 1.42x & 2.05 & 1.22x & 1.35x & 2.18 & 1.27x & 1.43x \\
      Ours    & \textbf{2.80} & \textbf{1.60x} & \textbf{2.06x} & \textbf{2.90} & \textbf{1.62x} & \textbf{2.13x} & \textbf{2.77} & \textbf{1.60x} & \textbf{2.05x} & \textbf{2.57} & \textbf{1.55x} & \textbf{1.91x} & \textbf{2.76} & \textbf{1.59x} & \textbf{2.04x} \\
      
      \bottomrule
    \end{tabular}
  \end{adjustbox}
\end{table*}

\subsection{Intermediate-Layer Visual State Bridging}  
Although visual input is masked during inference to mitigate noise, Table \ref{tab:visual_ablation_1} shows that the average accepted length decreases during training when visual information is either fully excluded or partially pruned. This indicates that even if the visual modality manifests as noise during inference, the explicit context it provides remains indispensable for parameter optimization. Consequently, effectively incorporating visual supervision during the training phase becomes pivotal.

Regarding the integration of visual information, MSD advocates for directly utilizing raw visual embeddings. Their core motivation is to preserve pure information unassimilated by the language modeling process, thereby preventing visual features from being polluted by textual semantics during deep interactions. However, this strategy overlooks the capacity bottleneck of the draft model. Constrained by its extremely shallow network structure, the model struggles to extract effective semantics from highly redundant and unaligned raw representations, which restricts its prediction accuracy.

To address this, we propose intermediate-layer visual state bridging(IVSB), which extracts the visual hidden state $h_{e_{vis}}^{m^*}$ from the interaction-active layer $m^*$ of the target model. As illustrated in Figure \ref{fig:architecture1} (b), this layer offers a dual advantage: on the one hand, it has achieved implicit semantic alignment and filtered out low-level noise; on the other hand, it avoids the loss of fine-grained details caused by over-compression in the final layers. This purified, high-quality feature is well-adapted to the limited processing capabilities of the draft model. Consequently, we set $m^*$ to half of the target model's total layers.

As illustrated in Figure \ref{fig:architecture_3}, We fuse this visual state with the text embedding $e_{txt}$ and the target model's second-to-last layer text state $h_{e_{txt}}^{(h)}$ (where superscript $h$ denotes the second-to-last layer of the target model, which has already internalized the preceding visual information). The initial input for the draft model, $z_{init}$, is formalized as:
\begin{equation}
    z_{init} = h_{e_{vis}}^{m^*} \oplus FC(e_{txt} \oplus h_{e_{txt}}^{h})
\end{equation}
By synergistically utilizing the explicitly purified visual features and the implicit visual context, the draft model avoids constructing complex low-level mappings, allowing it to focus on precise prediction based on well-adapted semantic representations.

To bridge the discrepancy between training and inference, we further introduce a multi-token prediction mechanism ~\cite{liu2024deepseek} (MTP). Given that during the inference phase, the draft model must rely on its own preceding outputs rather than the perfect states provided by the target model, we construct a recursive training pipeline. Specifically, starting from the initial input $z_{init}$, the draft model generates a textual hidden state $\hat{h}_{e_{txt}}$. Subsequently, this state is concatenated with the fixed intermediate visual anchor $h_{e_{vis}}^{m^*}$ to construct the recursive input $z_{rec}$:
\begin{equation}
    z_{rec} = h_{e_{vis}}^{m^*} \oplus FC(e_{txt} \oplus \hat{h}_{e_{txt}})
\end{equation}
This design compels the model to adapt to its self-generated distribution during training, effectively alleviating exposure bias. Crucially, the continuous reuse of the high-quality visual anchor within $z_{rec}$ provides a stable semantic reference for the autoregressive generation process.

\section{Experiment}

\paragraph{Hardware.} All evaluation experiments are conducted on 2x NVIDIA L20 and 2x NVIDIA A800 GPUs.

\paragraph{Tasks and Benchmarks.}We evaluate performance on video captioning and video description tasks, which typically require generating long-form text. The benchmarks employed include VideoDetailCaption ~\citep{vdc}, MVBench ~\citep{li2024mvbench}, LongVideoBench~\citep{wu2024longvideobench}, and VideoMME~\cite{fu2025video}.

\paragraph{Target Models}We select two widely used Vid-LLMs series: LLaVA-OneVision-7B~\citep{li2024llava} and Qwen2.5-VL-7B~\citep{ahmedqwen}.

\paragraph{Baselines} We include MSD~\cite{lin2025speculative}, ViSpec (orig.)~\cite{kang2025ViSpec}, ViSpec, and SpecVLM~\cite{ji2025specvlm} as our baselines. To ensure a fair comparison, except for ViSpec (orig.) and SpecVLM which utilize their officially released weights, all other baselines were reproduced and retrained with a batch size of 16. Note that ViSpec (orig.) adopts a batch size of 8. Further details are provided in Appendix ~\ref{sec:Implementation}.

\paragraph{Metrics} We assess acceleration performance using: (i) the speedup ratios of wall time and decoding latency relative to vanilla autoregressive decoding; and (ii) the average accepted length $\tau$, which directly reflects speculation accuracy. 

Given that our method is lossless, we do not evaluate output quality. With the exception of SpecVLM, which employs a static tree structure, our method and all other baselines utilize a dynamic tree structure. For the general setup, the dynamic configuration (denoted as total tokens-depth-width) is set to 30-4-8 on L20 GPUs (following ViSpec) and 48-5-10 on A800 GPUs. Specifically, to ensure a fair comparison in Figure 2, SpecVLM uses its default static setting (total budget 25, depth 5), while our method and other baselines are unified to a configuration of 25-5-8.

\begin{table}[t]
  \centering
  \caption{Performance results of Qwen2.5-VL-7B on video benchmarks on NVIDIA L20 GPUs. We report average accepted length ($\tau$) and decoding speedup ratio (DSR). Detailed visual token lengths: VideoDetailCaption ($\sim$17k), MVBench ($\sim$13k), LongVideoBench ($\sim$15k).}
  \label{tab:qwen_performance} 
  
  \begin{adjustbox}{max width=\columnwidth, center}
    \small
    \begin{tabular}{l cc cc cc cc}
      \toprule
      Method & \multicolumn{2}{c}{VideoDetailCaption} & \multicolumn{2}{c}{MVBench} & \multicolumn{2}{c}{LongVideoBench} & \multicolumn{2}{c}{Avg} \\
      
      \cmidrule(lr){2-3} \cmidrule(lr){4-5} \cmidrule(lr){6-7} \cmidrule(lr){8-9}
       & $\tau$ & DSR &$\tau$ & DSR & $\tau$ & DSR & $\tau$ & DSR \\
      \midrule
      
      \multicolumn{9}{c}{Temp = 0} \\
      \midrule
      
      MSD            & 1.11  & 0.52x & 1.86  & 0.95x & 1.75  & 0.88x & 1.57  & 0.78x \\
      SpecVLM        & \textbf{3.91}  & 1.20x & 3.80  & 1.09x & \textbf{3.82}  & 1.15x & 3.84  & 1.15x \\
      ViSpec (orig.) & 2.79  & 1.53x & 2.83  & 1.52x & 2.75  & 1.47x & 2.79  & 1.51x \\
      ViSpec         & 3.52  & 1.87x & 3.60  & 1.92x & 3.55  & 1.91x & 3.56  & 1.90x \\
      Ours  & 3.89 & \textbf{2.04x} & \textbf{3.87} & \textbf{2.06x} & 3.78 & \textbf{2.02x} & \textbf{3.85} & \textbf{2.04x} \\
      
      \midrule
      \multicolumn{9}{c}{Temp = 1} \\
      \midrule
      
      MSD            & 1.10  & 0.48x & 1.63  & 0.70x & 1.52  & 0.74x & 1.42  & 0.64x \\
      ViSpec         & 2.24  & 1.21x & 2.44  & 1.31x & 2.46  & 1.34x & 2.38  & 1.29x \\
      Ours  & \textbf{2.37} & \textbf{1.27x} & \textbf{2.55} & \textbf{1.36x} & \textbf{2.54} & \textbf{1.37x} & \textbf{2.49} & \textbf{1.33x} \\
      
      \bottomrule
    \end{tabular}
  \end{adjustbox}
\end{table}
\subsection{Main Results}
The experimental results in Table ~\ref{tab:video_performance} and Table ~\ref{tab:qwen_performance} demonstrate that our method establishes a significant performance advantage across all benchmarks when using LLaVA-OneVision-7B and Qwen2.5-VL-7B as target models. Specifically, MSD encounters severe performance bottlenecks when facing long sequences with an average visual length of 17k tokens. The attention dilution caused by full visual inputs makes it difficult to focus on core information, resulting in an average accepted length of only 1.11 on the VideoDetailedCaption task of Qwen2.5-VL-7B, causing a negative speedup of 0.52x. Although SpecVLM achieves an optimal average accepted length of 4.07 on LLaVA-OneVision by leveraging the target model reuse strategy, the heavy computational cost of the draft model severely offsets the gains from the high acceptance rate. Its decoding speedup of 1.41x is far lower than the 2.82x achieved by our method, making it difficult to meet real-time requirements. Furthermore, while ViSpec improves efficiency through visual token compression, its acceleration gains are limited by the difficulty in capturing the complex spatiotemporal details of long videos; its average speedup of 1.90x on Qwen2.5-VL-7B still lags behind the 2.04x achieved by our method.
\begin{table}[t]
  \centering
  \caption{Performance of Qwen2.5-VL-7B on VideoDetailCaption across varying visual token lengths, evaluated on A800 GPUs, in terms of average accepted length ($\tau$) and decoding speedup ratio (DSR).}
  \label{tab:performance_by_length} 
  
  \footnotesize
  \setlength{\tabcolsep}{0pt} 
  \begin{tabular*}{\columnwidth}{@{\extracolsep{\fill}} l c c c c c c c c c}
    \toprule
    \multicolumn{1}{c}{Method} & 
    \multicolumn{2}{c}{0.5k} & 
    \multicolumn{2}{c}{1.5k} & 
    \multicolumn{2}{c}{17k} & 
    \multicolumn{2}{c}{25k} \\
    \cmidrule(lr){2-3} \cmidrule(lr){4-5} \cmidrule(lr){6-7} \cmidrule(lr){8-9}
    & $\tau$ & DSR & $\tau$ & DSR & $\tau$ & DSR & $\tau$ & DSR \\
    \midrule
    
    MSD & 4.12 & 2.96x & 3.97 & 2.87x & 1.11 & 0.48x & 1.04 & 0.42x \\

    ViSpec (orig.) & 3.45 & 2.67x & 3.35 & 2.52x & 2.96 & 1.31x & 2.81 & 1.20x \\
    
    ViSpec & 4.29 & 3.22x & 4.18 & 3.10x & 3.74 & 1.64x & 3.42 & 1.48x \\

    MSD+VATA & 4.11 & 2.98x & 4.01 & 2.92x & 3.94 & 1.72x & 4.01 & 1.68x \\
    
    Ours & \textbf{4.51} & \textbf{3.30x} & \textbf{4.36} & \textbf{3.18x} & \textbf{4.32} & \textbf{1.86x} & \textbf{4.37} & \textbf{1.82x} \\
    
    \bottomrule
  \end{tabular*}
\end{table}

To validate effectiveness in real-world scenarios, Table~\ref{tab:performance_breakdown} presents a performance analysis of long-video inference. As visual inputs scale from 4.5K to 25K, prefill latency increases from 1.20s to 11.46s, causing its share of the total inference time to rise from 13.8\% to 38.7\%. Since speculative decoding cannot optimize the prefill phase, this increased overhead constrains the overall end-to-end speedup. Simultaneously, as shown in Table~\ref{tab:video_performance}, visual noise induced by long sequences degrades the performance of conventional draft models, causing the average accepted length of MSD to drop from 3.07 to 2.69; meanwhile, constrained by the computational burden of reusing the target model, SpecVLM sees its speedup decline from 1.41× to 1.24×. In contrast, our method demonstrates strong robustness, maintaining an average accepted length of 3.83 even with 25K visual tokens. Consequently, it achieves a 1.96× end-to-end speedup based on a 2.84× decoding speedup, verifying its effectiveness in long-sequence tasks.

\begin{table}[t]
    \centering
    \small
    \caption{Performance breakdown of LLaVA-OneVision-7B on VideoDetailedCaption across varying visual input lengths using L20 GPUs.}
    \label{tab:performance_breakdown}
    
    \begin{tabular}{l cccc}
        \toprule
        Visual Input Length  & 4.5K & 15K & 25K \\
        \midrule
        Average Accepted Length  & 3.07 & 2.78 & 2.69 \\
        Generated Tokens  & 418.86 & 450.53 & 455.10 \\
        Target Model Calls & 136.23 & 161.58 & 168.69 \\
        \midrule
        Wall (End-to-end) Time  (s)  & 8.71 & 19.28 & 29.59 \\
        Total Decode Time (s)  & 7.51 & 13.52 & 18.13 \\
        Prefill Time (s)  & 1.20 & 5.76 & 11.46 \\
        Prefill Ratio & 13.8\% & 29.9\% & 38.7\% \\
        Latency Per Decode Step (s)  & 0.055 & 0.083 & 0.11 \\
        \bottomrule
    \end{tabular}
\end{table}

Table~\ref{tab:performance_by_length} presents the acceleration performance evaluation across varying input lengths, compellingly demonstrating that our method combines efficiency in short-sequence scenarios with robustness in long-sequence contexts. Specifically, in the 0.5k short-sequence setting, we achieved a peak speedup of 3.30x and an average accepted length of 4.50, surpassing ViSpec's 3.22x and MSD's 2.96x, respectively. More critically, as the sequence length extends to 25k, traditional methods encounter severe challenges: MSD suffers a performance collapse with inference speed plummeting to 0.42x, while ViSpec's average accepted length drops to 3.42. In contrast, our method robustly maintains an accepted length of 4.35 and a speedup of 1.82x, fully demonstrating its consistent superiority across both short and long sequences.

\subsection{Ablation}

  
    
    

As shown in Table ~\ref{tab:ablation_study}, MTP training effectively mitigates distribution shifts and significantly enhances fundamental alignment capability, increasing the average effective length on the MME benchmark from 3.40 (Row 1) to 3.75 (Row 3). Building on this, introducing IVSB (Intermediate Visual Features) further exploits the model's potential, achieving a peak performance of 4.53 on short sequences (0.5k) in Row 5. However, constrained by noise accumulation from attention dilution, the performance of this configuration suffers a precipitous decline to 1.21 on 25k long sequences. The VATA mechanism effectively resolves this issue by offloading redundant visual computations, allowing our full method (Ours, Row 6) to stage a strong recovery to 4.37 at 25k. Notably, a comparison of Row 6 across different lengths demonstrates that VATA renders the model's performance nearly invariant to the negative impacts of increasing sequence length.
\begin{table}[t]
  \centering
  \caption{We conduct an ablation study on MTP, intermediate-layer visual state bridging(IVSB), and visually-aware text-anchored window attention(VATA) using Qwen2.5-VL-7B deployed on A800 GPUs as the base model on the MME and VideoDetailedCaption benchmarks, reporting the average accepted length across varying visual token lengths.}
  \label{tab:ablation_study} 
  
  \footnotesize
  \setlength{\tabcolsep}{0pt} 
  \begin{tabular*}{\columnwidth}{@{\extracolsep{\fill}} c c c c c c c c c c}
    \toprule
    \multicolumn{3}{c}{Method Components} & 
    \multicolumn{1}{c}{MME} & 
    \multicolumn{6}{c}{VideoDetailedCaption} \\
    \cmidrule(lr){1-3} \cmidrule(lr){4-4} \cmidrule(lr){5-10}
    MTP & IVSB & VATA & 0.4k & 0.5k & 1.5k & 6k & 17k & 25k \\
    \midrule
    
    $X$ & $X$ & $X$ & 3.40 & 3.95 & 3.87 & 3.29 & 2.97 & 2.95 \\
    $X$ & $X$ & $\checkmark$ & 3.41 & 3.95 & 3.85 & 3.83 & 3.77 & 3.86 \\
    $\checkmark$ & $X$ & $X$ & 3.75 & 4.44 & 4.31 & 3.87 & 3.58 & 3.51 \\
    $\checkmark$ & $X$ & $\checkmark$ & 3.75 & 4.45 & 4.34 & 4.24 & 4.24 & 4.29 \\
    $\checkmark$ & $\checkmark$ & $X$ & \textbf{3.82} & \textbf{4.53} & \textbf{4.38} & 3.08 & 1.91 & 1.21 \\
    $\checkmark$ & $\checkmark$ & $\checkmark$ & 3.79 & 4.51 & 4.36 & \textbf{4.29} & \textbf{4.32} & \textbf{4.37} \\
    
    \bottomrule
  \end{tabular*}
\end{table}

Experimental observations reveal that the cumulative noise associated with long sequences often masks the model's true capability limit; therefore, performance on short sequences serves as a superior indicator of Fundamental Capability. A comparison of configurations utilizing VATA (Rows 2, 4, and 6) demonstrates a clear correlation: the stronger the model's fundamental capability on short sequences, the more superior its long-sequence performance remains after applying VATA. This is particularly evident in Row 6, which achieves global optimality by leveraging the high baseline capability established by IVSB in conjunction with VATA. This finding validates the effectiveness of our synergistic design philosophy: prioritizing the enhancement of fundamental capability first, and then employing VATA to ensure long-sequence robustness.

\section{Conclusion}

In this paper, we delve into the root causes of speculative decoding inefficacy in long-video scenarios, identifying attention dilution and negative visual gain as the critical bottlenecks hindering lightweight draft models from handling ultra-long sequences. Based on the empirical validation of the visual semantic internalization mechanism within the target model, we propose the Sparrow framework. Through visual computational offloading via HSR-VATA, Sparrow not only effectively overcomes the capacity constraints of the draft model but also decouples visual processing costs from sequence length. Meanwhile, the synergistic integration of IVSB and the Multi-Token Prediction strategy further strengthens the predictive performance of the draft model. Benefiting from these designs, our method achieves a 2.82x decoding speedup on sequences of 25k visual tokens.

\section{Limitations}
While our method exhibits superior acceleration performance during the long-sequence decoding phase, the computational bottleneck in the prefill phase remains a primary limitation. As the visual input length scales, prefill latency increases substantially. Since speculative decoding inherently optimizes only the autoregressive generation process without addressing the prefill phase, this escalating overhead constrains the ceiling of the overall end-to-end speedup. Future work will aim to explore prefill acceleration techniques for long videos, such as incorporating visual token pruning. Although pruning strategies may introduce potential information loss, we believe that by designing refined importance-evaluation mechanisms, we can strike an optimal balance between inference speed and model accuracy while significantly reducing prefill costs, thereby achieving more efficient end-to-end inference.
\bibliography{custom}
\clearpage
\twocolumn  
\appendix

\section{Related Work}
\paragraph{Speculative Decoding}  
Speculative decoding has established itself as a pivotal acceleration paradigm in the realm of LLMs, significantly enhancing autoregressive generation efficiency while strictly adhering to the target model's original output distribution. Operating on a draft-and-verify mechanism, it employs a lightweight draft model to rapidly generate candidate token sequences, which are subsequently verified and corrected by the target model via a single parallel forward pass ~\cite{leviathan2023fast,chen2023accelerating}. Theoretically grounded in distribution consistency conditions ~\cite{leviathan2023fast}, this technique fundamentally ensures zero deviation in the inference process. In recent years, the field has witnessed continuous evolution with the emergence of variants such as tree-structured speculation ~\cite{li2024eagle,wang2025opt}, self-speculative decoding ~\cite{hooper2025speed,liu2024kangaroo,liu2024speculative}, and retrieval-augmented strategies ~\cite{wang2024speculative}. Complemented by system-level optimization frameworks ~\cite{sun2024triforce,monea2023pass,liu2024parallel,zhang2025dovetail}, these advancements continually push the boundaries of inference efficiency.

With the ubiquity of VLMs, speculative decoding has been extended to multimodal scenarios. Existing methodologies are primarily categorized into full visual input and compressed visual input strategies, depending on how the draft model handles the visual stream. Full visual input approaches ~\cite{huo2025spec,lin2025speculative,ganesan2025massv} typically reuse the target model's complete visual encoding pipeline. Conversely, compressed visual input approaches aim to reduce KV cache usage and latency by minimizing the scale of visual tokens. These methods generally follow two technical trajectories: (1) Token Aggregation and Mapping, exemplified by the adapter-based compression in ViSpec ~\cite{kang2025ViSpec}; and (2) Attention-Guided Dynamic Selection, which exploits attention sparsity to filter redundant tokens, with representative works including STD ~\cite{zhang2025sparse}, SPECVLM ~\cite{ji2025specvlm}, and DREAM ~\cite{hu2025dream}.

Despite differing perspectives, both our work and HiViS ~\cite{xie2025hivis} arrive at the same key insight: the draft model does not require explicit visual tokens and can effectively reason using only the text-side hidden states of the target model, in which visual semantics are implicitly encoded. The key difference lies in their usage: HiViS completely discards visual tokens during both training and inference, whereas we retain them during training to provide essential visual supervision and only mask them during inference. Moreover, HiViS targets short-sequence tasks to reduce redundancy, while our method addresses long-video scenarios, mitigating the visual negative gain caused by attention dilution in ultra-long visual sequences.

\paragraph{Stage-wise Visual Processing in VLMs}  

Research on the visual processing mechanisms of VLMs has evolved from phenomenological observation to mechanistic elucidation. Early works, such as FastV ~\cite{chen2024image}, uncovered the non-uniformity of vision-text fusion based on attention distributions, highlighting that visual weights in deep layers are negligible (merely 0.21\%) and proposing single-stage filtering to eliminate redundancy. With the deepening of mechanistic research, ~\cite{zhang2025cross} and ~\cite{yin2025lifting} utilized attention knockout experiments to reveal the law of Layer-wise Functional Differentiation in cross-modal fusion: shallow layers primarily inject global features, middle layers serve as the core window for fine-grained semantic alignment, and deep layers govern final prediction. Building on this, HiMAP ~\cite{yin2025lifting} introduced a dynamic pruning strategy specifically targeting weak interactions in shallow layers and high aggregation in deep layers. Recent studies ~\cite{fan2025visipruner,zhang2025vscan}have further synthesized this into a three-stage framework: shallow layers function mainly as "Attention Sinks" or focus on text processing; middle layers act as the pivotal window for cross-modal fusion; and deep layers are dedicated to linguistic structure optimization. Despite minor discrepancies regarding the specific role of shallow layers, a strong consensus has emerged: the substantive contribution of visual information is highly concentrated in the intermediate layers. Once the multimodal representation is constructed in the middle layers, raw visual tokens in the deep layers become redundant due to semantic saturation.

\begin{table}[t]
    \centering
    \caption{Impact of visual information injection on draft model performance during inference.}
    
    \label{tab:visual_ablation}
    \begin{tabular*}{\linewidth}{@{\extracolsep{\fill}}l c c}
        \toprule
        Draft Model Input & ChartQA & AI2D \\
        \midrule
        Visual + Text         & 4.25 & 4.28 \\
        50\% Visual + Text    & 4.26 & 4.30 \\
        12.5\% Visual + Text  & 4.25 & 4.30 \\
        0\% Visual + Text     & 4.27 & 4.31 \\
        \bottomrule
    \end{tabular*}
\end{table}

\section{Task and Benchamark Details}

Tasks and Benchmark Details. Given that Sparrow is primarily dedicated to achieving lossless acceleration during the decoding stage, we select video captioning and video description as our core evaluation tasks. These tasks require the model to deeply understand specific subjects and actions within the video and generate long textual descriptions rich in visual information. Unlike conventional Video Question Answering (VQA), which typically involves direct answer selection, these long-form generation tasks better highlight the acceleration advantages of the model in long-sequence decoding.

For the datasets in our main experiments, to ensure comprehensive coverage of varying video durations and application scenarios, we randomly sampled 100 instances for performance evaluation from mainstream benchmarks, including VideoDetailCaption ~\citep{vdc}, MVBench ~\citep{li2024mvbench}, LongVideoBench~\citep{wu2024longvideobench}, and VideoMME~\cite{fu2025video}.

Furthermore, prior to the main experiments, to verify the universality of the phenomena observed in this work across different tasks and modalities, we also conducted sampling analysis on VideoChatGPT ~\citep{vcg} and VideoDetailDescription ~\citep{Maaz2023VideoChatGPT}, as well as the visual question answering task AI2D ~\citep{kembhavi2016diagram}, the document and chart understanding benchmark ChartQA ~\citep{masry2022chartqa}, and the comprehensive capability benchmark MME ~\citep{fu2025mme}.

\section{Implementation Details}
\label{sec:Implementation}
SpecVLM adopts a self-speculative strategy, wherein the draft model directly reuses the weights of the target model. By aggressively pruning redundant visual tokens at the input level, it significantly reduces KV cache occupancy and computational overhead. This input-level sparsification enables inference acceleration without altering the underlying model structure. In contrast, MSD, Vispec, and our proposed method opt to construct an independent single-layer draft model. Architecturally, this model mirrors a single decoder layer of the target model, aiming to maximize speculative efficiency through extreme structural lightweighting while ensuring feature space alignment and compatibility.

\subsection{Visual Information Processing Strategies in Draft Models}
SpecVLM leverages the draft model's insensitivity to missing visual information by proposing a verifier-guided two-stage video token pruning strategy. Guided by the attention signals of the target model (the verifier), this method prioritizes retaining highly informative tokens while performing spatially uniform pruning on low-attention redundant tokens. Ultimately, this approach reduces the visual input for the draft model by approximately 90\%.

MSD posits that visual and textual characteristics are distinct, thus adopting an architectural decoupling approach by directly using raw visual embeddings as input. Unlike features processed by the LLM, these raw embeddings remain uncontaminated by the language modeling process. Consequently, they preserve the most pristine visual information, providing more accurate guidance for the draft model.

ViSpec efficiently processes visual information through a dual fusion mechanism. First, it utilizes a lightweight visual adapter to compress redundant image tokens into compact representations and integrates them into the draft model's attention layers, thereby retaining key spatial information while reducing computational overhead. Second, it extracts a global image feature vector and injects it into every subsequently generated text token. This provides the model with continuous and robust global visual context, significantly enhancing the accuracy of multimodal predictions.

\subsection{Selection of Training Data}
Building on the analysis in Sec.~\ref{sec:Insight}, directly transferring a small model trained solely on image-text pairs to long-video scenarios inevitably leads to a double collapse. The excessively long visual sequences (>10k tokens) not only exceed the pre-trained positional encoding window but also introduce significant background noise. This distracts the draft model, causing critical fused representations to be submerged amidst the noise. However, a turning point emerged when applying the VATA strategy—which physically masks raw visual inputs to force reliance on hidden states. We surprisingly observed that the average acceptance length no longer decays as video token length increases. This phenomenon provides decisive empirical support for our hypothesis: once deep fused hidden states are available, raw visual inputs become purely redundant noise during inference.

Consequently, we adopt a minimalist data strategy, eliminating the need for expensive and scarce long-video datasets for draft model training. The rationale lies in the HSR strategy, which effectively offloads the complex computations of visual perception, spatiotemporal modeling, and cross-modal alignment to the Target Model. Whether the input is a single-frame image or a long video, the draft model's task is unified and simplified to merely decoding the deep fused representations from the Target Model. Therefore, we train exclusively using low-cost Image-Text Pairs. This approach not only circumvents the massive computational overhead of processing contexts exceeding 20k tokens but also enables the small model to focus solely on mapping deep fused features to the output vocabulary. This facilitates a seamless transfer of capabilities learned from image-text data to long-video scenarios, achieving robust cross-modal generalization with minimal resource costs.
\begin{figure}[t]
  \centering 
  \includegraphics[width=\linewidth]{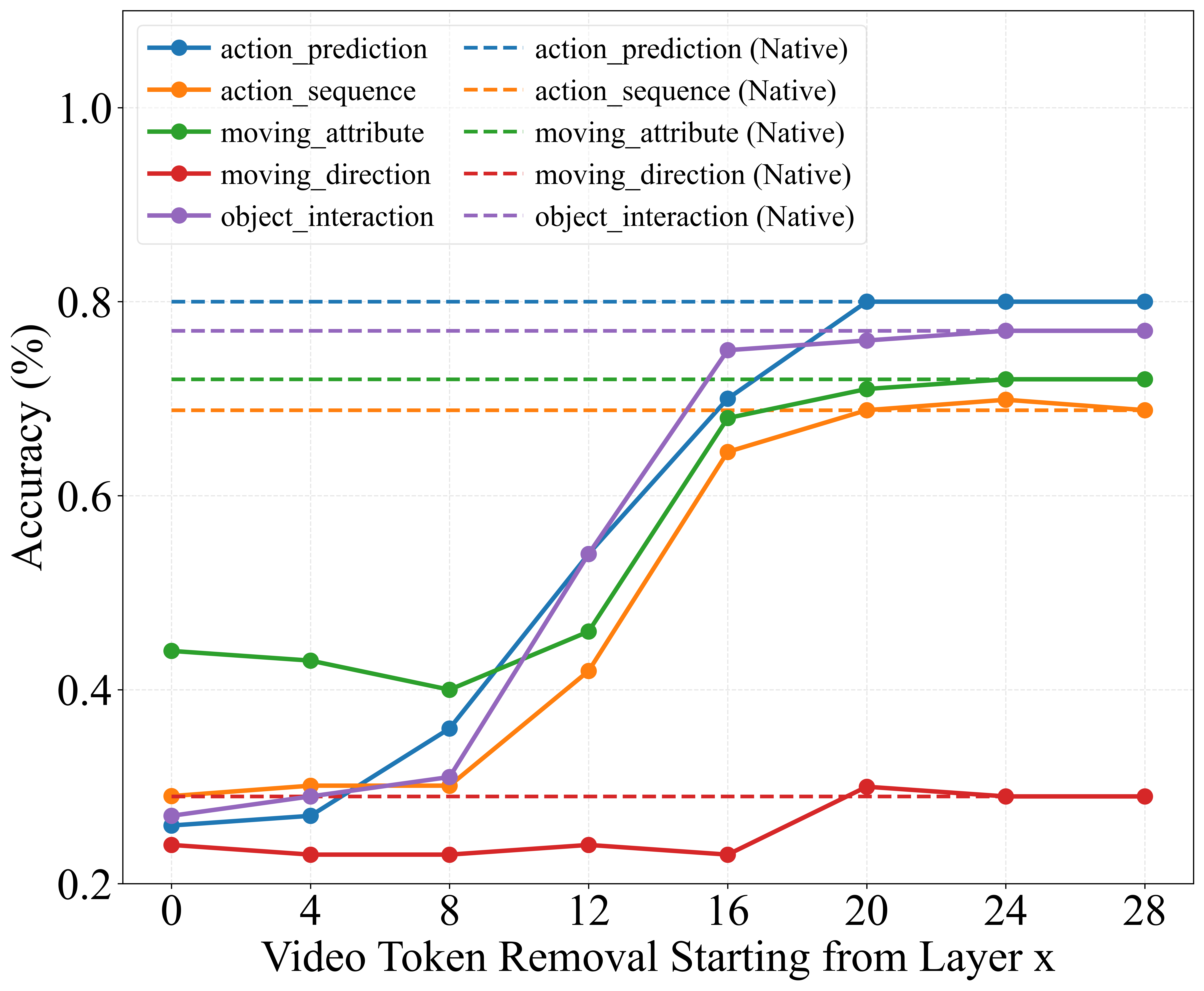}
  
  \caption{Layer-wise importance analysis of visual tokens in LLaVA-OneVision-7B. The figure compares task accuracy after removing visual tokens starting from layer x (solid lines) with the native baseline performance (dashed lines).
}
  \label{fig:architecturexxx}
\end{figure}

\subsection{Datasets and Training Details}
Following the data construction paradigm of ViSpec, we established a two-stage training dataset. In the first phase (Text Alignment), we randomly sampled 68,000 multi-turn conversation samples from the ShareGPT dataset to ensure a robust linguistic foundation. The second phase (Multimodal Fine-tuning) utilized the LLaVA Visual Instruct Pretrain LCS dataset ~\cite{li2024llava-a}, with an equal sample size of 68,000. Crucially, we implemented an instruction-driven data augmentation strategy during this phase. By appending the command "Please answer with at least 1000 words" to the original prompts, we induced the target VLM to generate long-form synthetic responses enriched with detailed explanations and complex reasoning chains, thereby constructing a high-quality dataset for multimodal instruction fine-tuning.

MSD adopts a two-stage training strategy spanning a total of 40 epochs. The first stage utilizes pure text data, while the second stage employs a linear dynamic mixing strategy that progressively decreases the proportion of text data and increases the visual data as training proceeds. In contrast, both ViSpec and our method follow a different two-stage paradigm: the first stage consists of text-only training, followed by a direct switch to exclusive multimodal data training in the second stage. For these methods, each stage is trained for 20 epochs. All methods were trained with a batch size of 16.
\begin{figure}[t]
  \centering 
  \includegraphics[width=\linewidth]{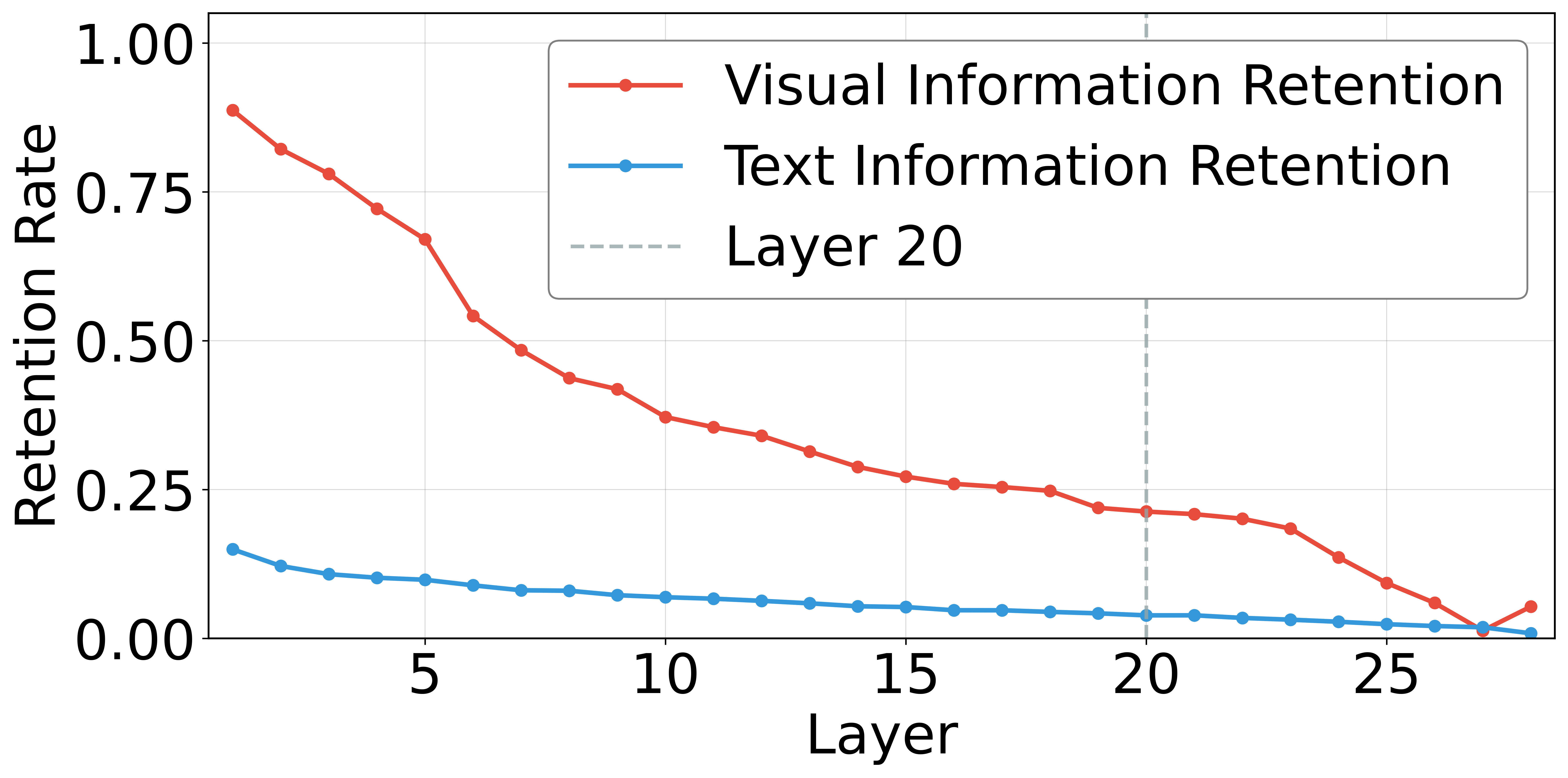}
  
  \caption{Layer-wise information retention analysis of visual and text modalities in Qwen2.5-VL-7B. The curves illustrate the retention rates calculated based on similarity, with red denoting visual features and blue denoting text features. The vertical dashed line marks the critical Layer 20.
}
  \label{fig:architecture1xxx}
\end{figure}
\section{Supplementary Insights and Analysis}
To further elucidate the phenomenon observed in the main text—where removing visual information after Layer 20 does not compromise performance—we introduce the information retention rate as a supplementary perspective, specifically defined as the average cosine similarity between the hidden states at layer l and the original input embeddings. As shown in Figure ~\ref{fig:architecture1xxx}, the visual retention rate exhibits a precipitous decline in the shallow and middle layers, standing in sharp contrast to the relatively flat text curve. This indicates that visual features undergo intense transformation and reconstruction during the early stages to align with the text modality. More critically, the similarity score drops below 0.25 near Layer 20, corresponding precisely with the results of the truncation experiment. Consequently, 0.25 can be regarded as a critical Empirical Threshold: it signifies the completion of visual information fusion and provides quantitative guidance for future Visual Pruning strategies—suggesting that once the similarity falls below this threshold, visual tokens can be safely removed without causing performance degradation.

We present the layer-wise truncation experiment based on the LLaVA-OneVision-7B model. As shown in Figure ~\ref{fig:architecturexxx}, the results reproduce the phenomenon observed in the main text: while explicit visual signals are indispensable in the early network stages, prediction performance remains robust beyond the 20th layer, even after completely removing the visual flow and relying solely on text hidden states.

\section{Extension to Image Tasks and Short-Sequence Generalization}

Table ~\ref{tab:video_performance} and Table ~\ref{tab:qwen_performance} have demonstrated the superiority of our method in long-video understanding tasks. To evaluate its generalization capability, we conduct extended experiments on image-based tasks and compare it against efficient inference methods specifically designed for images or short-sequence visual inputs, including MSD, ViSpec, and ViSpec (orig.). The results show that our method maintains high inference efficiency while exhibiting stronger generality.

\begin{table*}
  \centering
  \caption{Performance results of Qwen2.5-VL-7B on benchmarks on NVIDIA GPUs. We report average accepted length ($\tau$), decoding speed (Tokens/Sec), and speedup ratio (ESR).}
  \label{tab:performance_1}   
  
  \begin{adjustbox}{max width=\linewidth, center}
    \small
    \begin{tabular}{l *{4}{ccc} cc}
      \toprule
      Method & \multicolumn{3}{c}{COCO caps} & \multicolumn{3}{c}{MME} & \multicolumn{3}{c}{SQA} & \multicolumn{3}{c}{MM-Vet} & \multicolumn{2}{c}{Avg.} \\
      
      \cmidrule(lr){2-4} \cmidrule(lr){5-7} \cmidrule(lr){8-10} \cmidrule(lr){11-13} \cmidrule(lr){14-15}
       & $\tau$ & Tokens/Sec & ESR & $\tau$ & Tokens/Sec & ESR & $\tau$ & Tokens/Sec & ESR & $\tau$ & Tokens/Sec & ESR & $\tau$ & ESR \\
      \midrule
      
      \multicolumn{15}{c}{T = 0} \\
      \midrule
      
      MSD            & 3.77 & 79.63 & 2.00x      & 3.49 & 68.37 & 1.83x & 3.75 & 76.66 & 1.95x & 3.60 & 70.14 & 1.87x & 3.65 & 1.91x \\
      ViSpec(orig.) & 3.37 & 72.94 & 1.83x   & 3.19 & 64.85 & 1.71x & 3.26 & 68.73 & 1.75x & 3.20 & 64.81 & 1.72x & 3.26 & 1.75x \\
      ViSpec         & 4.00  & 84.93 & 2.13x   & 3.62 & 71.44 & 1.88x & 3.78 & 78.01 & 1.98x & 3.72 & 72.83 & 1.94x & 3.78 & 1.98x \\
      Ours - VATA    & \textbf{4.13} & 85.61 & \textbf{2.15x}   & \textbf{3.78} & \textbf{72.59} & \textbf{1.92x} & \textbf{3.86} & 77.50  & 1.97x & \textbf{3.82} & 73.42 & 1.95x & \textbf{3.90} & \textbf{2.00x} \\
      Ours           & 4.11 & \textbf{85.80} & \textbf{2.15x}   & 3.73 & 72.17 & 1.90x & 3.84 & \textbf{78.19} & \textbf{1.99x} & 3.80 & \textbf{74.16} & \textbf{1.97x} & 3.87 & \textbf{2.00x} \\
      
      \midrule
      \multicolumn{15}{c}{T = 1} \\
      \midrule
      
      MSD            & 2.85 & 61.83 & 1.55x   & 2.73 & 56.78 & 1.49x & 3.24 & 67.30 & 1.71x & 2.94 & 62.7  & 1.66x & 2.94 & 1.60x \\
      ViSpec(orig.) & 2.68 & 59.63 & 1.50x   & 2.52 & 53.90 & 1.42x & 2.85 & 61.61 & 1.56x & 2.70 & 58.25 & 1.55x & 2.69 & 1.51x \\
      ViSpec         & 2.94 & 65.18 & \textbf{1.64x}   & 2.78 & 58.84 & 1.55x & 3.21 & 68.00 & 1.72x & 2.91 & 63.52 & 1.69x & 2.96 & 1.65x \\
      Ours - VATA    & \textbf{3.00} & 64.82 & 1.63x   & \textbf{2.95} & \textbf{59.84} & \textbf{1.57x} & \textbf{3.23} & 67.72 & 1.72x & 2.80 & 58.86 & 1.56x & 3.00 & 1.62x \\
      Ours           & 2.99 & \textbf{65.44} & \textbf{1.64x}   & 2.84 & 59.11 & 1.55x & \textbf{3.23} & \textbf{68.58} & \textbf{1.74x} & \textbf{3.00} & \textbf{64.02} & \textbf{1.70x} & \textbf{3.02} & \textbf{1.66x} \\
      
      \bottomrule
    \end{tabular}
  \end{adjustbox}
\end{table*}

As shown in Table ~\ref{tab:performance_1}, our method achieves the best average speedup ratio across four image-related benchmark tasks. The experiments were conducted on a single NVIDIA virtual GPU (vGPU-32GB) rented from AutoDL, which provides up to 32 GB of video memory, 52.22 TFLOPS of FP32 compute performance, and approximately 103 Tensor TFLOPS of FP16 compute performance, paired with an Intel(R) Xeon(R) Gold 6459C CPU. Further experiments reveal that when the VATA mechanism is disabled (Ours – VATA), the draft model explicitly processes all visual tokens, yielding the highest average accepted length. This is attributed to the model’s ability to effectively attend to key visual and textual cues in short-sequence scenarios, avoiding the attention dilution commonly observed in long-video settings due to an overabundance of visual frames.
When VATA is enabled (Ours), visual features are entirely skipped during draft generation. Although this design reduces the average accepted length, it further shortens the draft model’s inference latency. On COCO Captions ~\cite{chen2015microsoft}(COCO Caps), ScienceQA ~\cite{lu2022learn} (SQA), and MM-Vet ~\cite{yu2023mm}, the time saved during drafting sufficiently compensates for the additional decoding steps incurred by the lower acceptance length, resulting in an overall speedup. In contrast, on the MME ~\cite{fu2025mme} benchmark, the reduction in drafting time does not offset the cost of extra decoding steps, leading to a slight slowdown.
Notably, in multimodal short-sequence settings, explicitly removing visual inputs does not significantly degrade performance, owing to the visual semantic internalization phenomenon described in Section ~\ref{sec:visual_internalization31}. Our method consistently outperforms ViSpec and MSD across all tasks, demonstrating robustness and broad applicability across different modalities and task configurations.

\end{document}